\documentclass[10pt,twocolumn,letterpaper]{article}
 
\usepackage[pagenumbers]{cvpr} 

\usepackage[dvipsnames]{xcolor}

\definecolor{cvprblue}{rgb}{0.21,0.49,0.74}
\usepackage[pagebackref,breaklinks,colorlinks,citecolor=cvprblue]{hyperref}
\usepackage{breqn}
\usepackage{bm}
\usepackage{caption}

\newcommand{\topic}[1]
{
\vspace{1mm}\noindent\textbf{#1}
}

\title{TextureDreamer: Image-guided Texture Synthesis  through Geometry-aware Diffusion}

\author{Yu-Ying Yeh$^{13}$ \quad
Jia-Bin Huang$^{23}$ \quad
Changil Kim$^3$ \quad
Lei Xiao$^3$ \quad
Thu Nguyen-Phuoc$^3$ \quad
Numair Khan$^3$ \quad \\
Cheng Zhang$^3$ \quad
Manmohan Chandraker$^1$ \quad
Carl S Marshall$^3$ \quad
Zhao Dong$^3$ \quad
Zhengqin Li$^3$ \quad
\and
$^1$University of California, San Diego
\and
$^2$University of Maryland, College Park
\and
$^3$Meta
}

\begin{document}

\twocolumn[{
    \renewcommand\twocolumn[1][]{#1}
    \maketitle
    
  \begin{center}

  \includegraphics[width=1.0\linewidth]{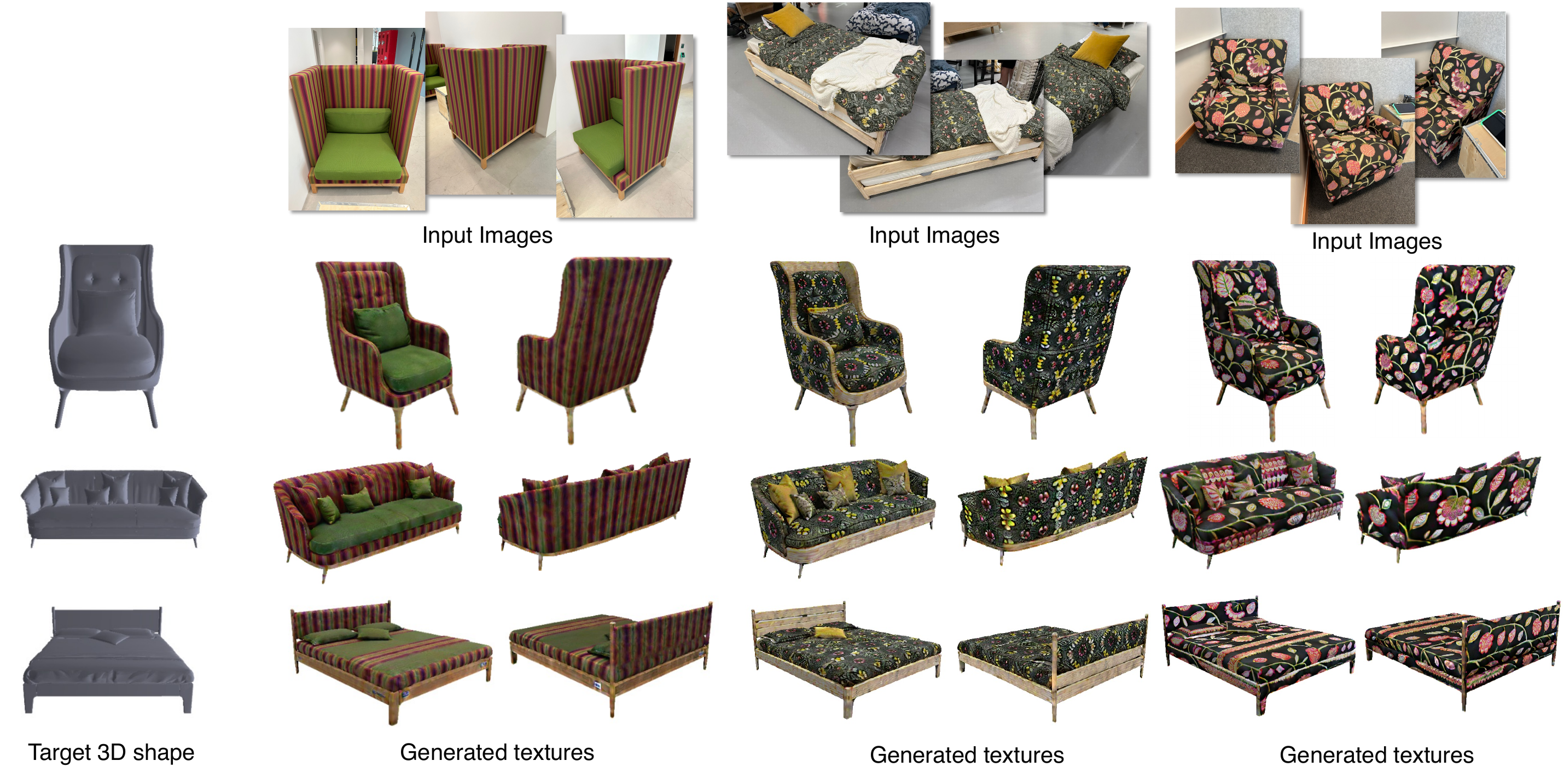}

  \vspace{-3mm}

\captionof{figure}{
\textbf{Texture transfer from sparse images.} Given a small number of images and a target mesh, our method synthesizes geometry-aware texture that looks similar to the input appearances for diverse objects. 
}
  \label{fig:teaser}
  \end{center}

    }]

\begin{abstract}
\vspace{-3mm}
We present TextureDreamer, a novel image-guided texture synthesis method to transfer relightable textures from a small number of input images (3 to 5) to target 3D shapes across arbitrary categories. 
Texture creation is a pivotal challenge in vision and graphics. 
Industrial companies hire experienced artists to manually craft textures for 3D assets.
Classical methods require densely sampled views and accurately aligned geometry, while learning-based methods are confined to category-specific shapes within the dataset. 
In contrast, TextureDreamer can transfer highly detailed, intricate textures from real-world environments to arbitrary objects with only a few casually captured images, potentially significantly democratizing texture creation. 
Our core idea, \textit{personalized geometry-aware score distillation (PGSD)}, draws inspiration from recent advancements in diffuse models, including personalized modeling for texture information extraction, variational score distillation for detailed appearance synthesis, and explicit geometry guidance with ControlNet. Our integration and several essential modifications substantially improve the texture quality. Experiments on real images spanning different categories show that TextureDreamer can successfully transfer highly realistic, semantic meaningful texture to arbitrary objects, surpassing the visual quality of previous state-of-the-art. Project page: \url{https://texturedreamer.github.io}
\end{abstract}    

\section{Introduction}
\label{sec:intro}
High-quality 3D content is indispensable for a wide range of critical applications, including AR/VR, robotics, film, and gaming. 
In recent years, remarkable progress has been made in democratizing 3D content creation pipelines, facilitated by advancements in 3D reconstruction \cite{mildenhall2021nerf,muller2022instant} and generative models \cite{goodfellow2020generative,sohl2015deep}. 
While substantial attention has been devoted to exploring the \emph{geometry component} \cite{wu2016learning,choy20163d,chang2015shapenet} and neural implicit representations \cite{park2019deepsdf}, such as NeRF \cite{mildenhall2021nerf}, creation of high-quality \emph{textures} is relatively under-explored. 
Textures are pivotal in creating realistic, highly detailed appearances and are integral to various graphics pipelines, where industry has traditionally relied on professional, experienced artists to craft textures. 
This process usually involves manually authoring procedural graphs \cite{substance} and UV maps, making it expensive and inefficient. 
Automatically transferring the diverse visual appearance of objects around us to the texture of any target geometry would thus be highly beneficial.

\begin{figure}[t]
  \centering
  \includegraphics[width=\linewidth]{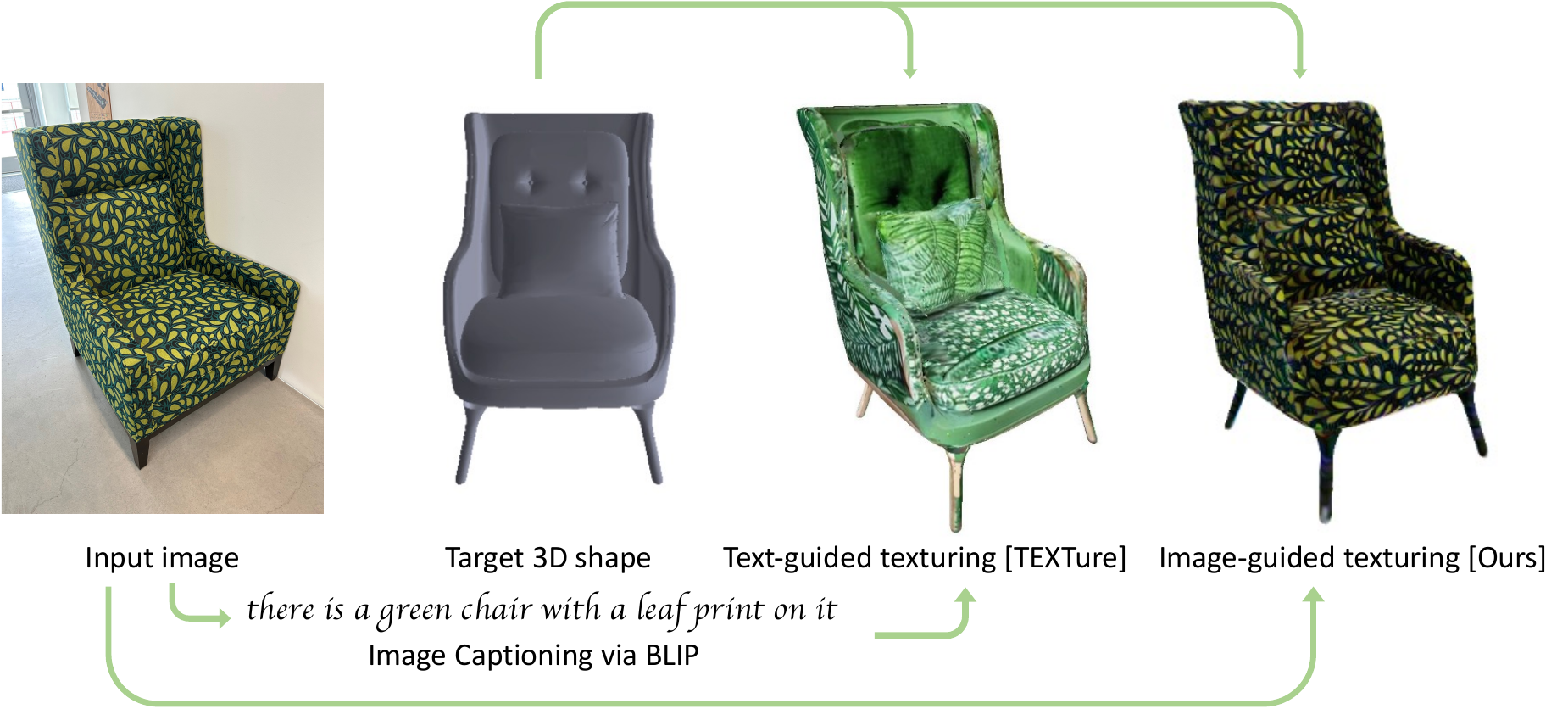}

  \caption{\textbf{Limitation of text-guided texturing.} Compared to text-guided texturing method which requires a captioning method to generate a text prompt which might not express all the details of the image, image-based guided texturing can be more effective and more expressive. Image captioning is predicted by BLIP~\cite{li2022blip}, text-guided texturing is generated via TEXTure~\cite{richardson2023texture}, and image-guided result is from our method.}
  \label{fig:text_vs_img}
\end{figure}

We present \emph{TextureDreamer}, a novel framework to create high-quality relightable textures from sparse images. 
Given 3 to 5 randomly sampled views of an object, we can transfer its texture to an target geometry that may come from a different category. 
This is an extremely challenging problem, as previous texture creation methods usually either require densely sampled views with aligned geometry \cite{bi2017patch,zhou2014color,levoy2000digital}, or can only work for category-specific shapes \cite{siddiqui2022texturify,bokhovkin2023mesh2tex,pavllo2021learning,henderson2020leveraging}. 
Our framework draws inspiration from recent advancements in diffusion-based generative models \cite{sohl2015deep,song2020score,ho2020denoising}. 
Trained on billions of text-image pairs, these diffusion models enable text-guided image generation with extraordinary visual quality and diversity \cite{ramesh2022hierarchical}. 
Pioneering works have applied these pre-trained 2D diffusion models to text-guided 3D content creation \cite{poole2022dreamfusion,lin2023magic3d,wang2023prolificdreamer}.
However, a common limitation among those methods is that \emph{text-only input} may not be sufficiently expressive to describe complex, detailed patterns, as demonstrated in Figure~\ref{fig:text_vs_img}. 
In contrast to text-guided methods, we effectively extract texture information from a small set of input images by fine-tuning the pre-trained diffusion model with a unique text token \cite{gal2022image,ruiz2023dreambooth}. 
Our framework, therefore, addresses the challenge of accurately describing complex textures.

The Score Distillation Sampling (SDS) \cite{poole2022dreamfusion,wang2023score} is one core element that bridges pre-trained 2D diffusion models with 3D content creation. 
It is widely used to generate and edit 3D contents by minimizing the discrepancy between the distribution of rendered images and the distribution defined by the pre-trained diffusion models~\cite{lin2023magic3d,metzer2023latent}. 
Despite its popularity, two well-known limitations impede its ability to generate high-quality textures. 
First, it tends to create over-smoothed and saturated appearances due to the unusually high classifier-free guidance necessary for the method to converge. 
Second, it lacks the knowledge to generate a 3D-consistent appearance, often resulting in multi-face artifacts and mismatches between textures and geometry.

We propose two key design choices to tackle these challenges.
Instead of using SDS, we build upon Variational Score Distillation (VSD) in our optimization approach, which can generate much more photorealistic and diverse textures. 
Initially introduced in ProlificDreamer \cite{wang2023prolificdreamer}, VSD treats the whole 3D representation as a random variable and aligns its distribution with the pre-trained diffusion model. 
It does not need a large classifier-free guidance weight to converge, which is essential to create a realistic and diverse appearance. 
However, na\"{i}vely applying VSD update does not suffice for generating high-quality textures in our application. 
We identify a simple modification that can improve texture quality while slightly reducing the computational cost.
Additionally, VSD loss alone cannot fully solve the 3D consistency issue. 
Fine-tuning on sparse inputs makes converging harder, as observed by previous work \cite{raj2023dreambooth3d}. 
We, therefore, explicitly condition our texture generation process on geometry information extracted from the given mesh by injecting rendered normal maps into the fine-tuned diffusion model through the ControlNet \cite{zhang2023adding} architecture. 
Our framework, designated as personalized geometry aware score distillation (PGSD), can effectively transfer highly detailed textures to diverse geometry in a semantically meaningful and visually appealing manner.
Extensive qualitative and quantitative experiments demonstrate that our framework substantially outperforms state-of-the-art texture-transfer methods.

\section{Related Works}
\label{sec:related}

\topic{Texture synthesis and reconstruction}
Classical texture creation methods involve sampling from a distribution derived from the neighborhood \cite{efros1999texture,kopf2007solid}, tiling repetitive patterns \cite{kwatra2003graphcut} or fusing multi-view images onto the object surfaces \cite{bi2017patch,zhou2014color,levoy2000digital}. 
The former two fall short in creating semantic meaningful textures while the latter one requires highly accurate geometry reconstruction. 
Numerous learning-based methods were proposed to learn texture creation from large-scale 3D datasets \cite{chen2022auv,bokhovkin2023mesh2tex,siddiqui2022texturify,henderson2020leveraging,pavllo2021learning} but are confined to specific categories within the dataset.
Recent works also use CLIP model \cite{radford2021learning} for text-guided texture generation of arbitrary objects \cite{michel2022text2mesh,lei2022tango,mohammad2022clip,ma2023x}, but their texture qualities are usually low. 
In contrast, TextureDreamer can create semantically meaningful, high-quality textures for arbitrary objects using uncorrelated sparse images. 
Traditionally, textures are represented as a 2D image and projected to object surfaces through UV mapping. 
Leveraging the recent progress in neural implicit representation, our method, along with recent developments in inverse rendering \cite{gao2022get3d,chan2022efficient,cai2022physics,sun2023neural} and 3D generation \cite{gao2022get3d,chan2022efficient}, represents texture as a neural implicit texture field.

\topic{Diffusion models}
Diffusion models \cite{sohl2015deep} have emerged as the state-of-the-art generative models \cite{ho2020denoising,song2020score}, demonstrating exceptional visual quality \cite{ramesh2022hierarchical}. 
Its training and inference involve iteratively adding noise with different variances and denoise the data. 
Trained on internet-scale image-text pair datasets \cite{ramesh2022hierarchical}, these pre-trained models exhibit unprecedented capability in text-guided image synthesis and have proven successful in various image editing tasks. 
Recent works also manage to fine-tune pre-trained diffusion models on much smaller datasets or even a few images to facilitate customized/personalized image synthesis \cite{ruiz2023dreambooth} and image generation conditioned on multi-modal data \cite{zhang2023adding}, such as normal and semantic maps.
Building upon this progress, TextureDreamer can effectively extract texture information from sparse views and transfer it to a novel target object in a geometry-aware manner. 

\topic{3D generation with 2D diffusion priors}
Diffusion-based 3D content creation has very recently gained substantial interest. 
Several methods directly train 3D diffusion models to generate 3D content in various representations, including point cloud \cite{luo2021diffusion}, neural radiance filed \cite{karnewar2023holodiffusion}, hyper-network \cite{erkocc2023hyperdiffusion} and texture \cite{yu2023texture}. 
Others utilize pre-trained 2D diffusion models by either progressively fusing generated images from different views \cite{richardson2023texture,cao2023texfusion,chen2023text2tex,albahar2023single} or optimizing the 3D representation through score distillation sampling \cite{lin2023magic3d,metzer2023latent,poole2022dreamfusion} and its improved variations \cite{katzir2023noise,wang2023prolificdreamer}.
While many methods concentrate on text-guided 3D generation, fewer attempt to leverage diffusion models to generate 3D content from images. 
A number of concurrent works fine-tune 2D diffusion models on large-scale 3D datasets for sparse view reconstruction \cite{qian2023magic123,shi2023mvdream}, primarily focusing on whole 3D object reconstruction. 
In contrast, TextureDreamer targets transferring textures from a small number of images to a target 3D shape with unmatched geometry. 
Dreambooth3D \cite{raj2023dreambooth3d} and TEXTure \cite{richardson2023texture} extract information from sparse views into a new text token and fine-tuned diffusion model weights, which can be used to generate personalized 3D object or texture unseen objects.
TextureDreamer employs a similar method to extract information from sparse images. 
However, it differs from prior works on utilizing the extracted information for texture generation, leading to improvements in consistency and photorealism. 

\section{Method}

\begin{figure*}[t]
  \centering
  \includegraphics[width=1.0\linewidth]{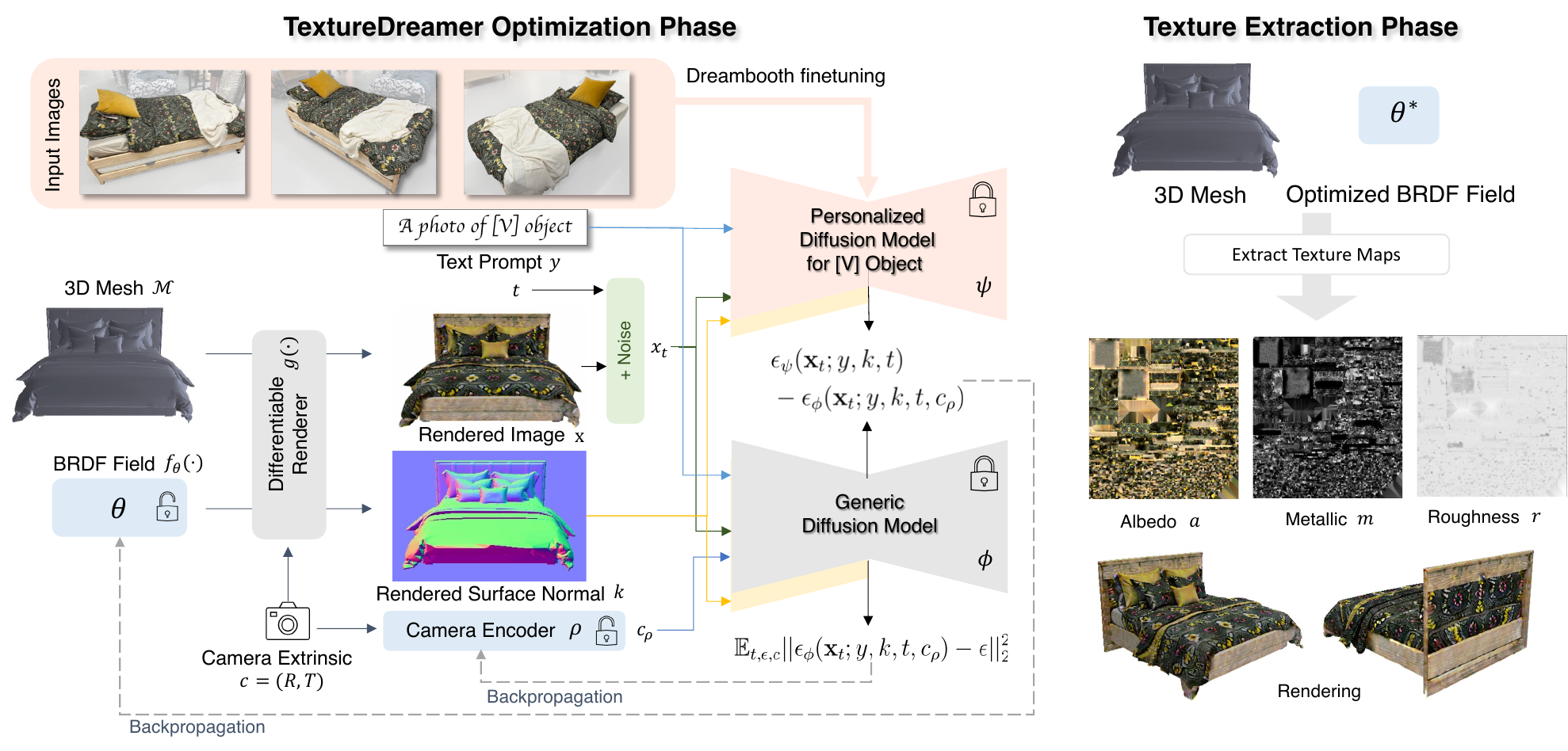}
  \vspace{-3mm}

  \caption{\textbf{Overview of TextureDreamer}, 
  a framework which synthesizes texture for a given mesh with appearance similar to 3-5 input images of an object.
  We first obtain personalized diffusion model $\psi$ with Dreambooth~\cite{ruiz2023dreambooth} finetuning on input images. 
  The spatially-varying bidirectional reflectance distribution (BRDF) field $f_{\theta}$ for the 3D mesh $\mathcal{M}$ is then optimized through personalized geometric-aware score distillation (PGSD) (detailed in Section~\ref{sec:pgsd}).
  After optimization finished, high-resolution texture maps corresponding to albedo, metallic, and roughness can be extracted from the optimized BRDF field.}
  \label{fig:method}
\end{figure*}
We propose TextureDreamer, a framework which synthesizes geometry-aware texture for a given mesh with appearance similar to 3-5 input images of an object.
In Section~\ref{sec:preliminary}, we first introduce preliminaries on Dreambooth \cite{ruiz2023dreambooth}, ControlNet \cite{zhang2023adding} and score distillation sampling \cite{poole2022dreamfusion,wang2023score,wang2023prolificdreamer}. 
In Section~\ref{sec:pgsd}, we propose personalized geometry-aware score distillation (PGSD), which is our core technical contribution that enables high-quality image-guided texture transfer from sparse images to arbitrary geometries.

\subsection{Preliminaries} 
\label{sec:preliminary}

\topic{Dreambooth}~\cite{ruiz2023dreambooth} is a simple yet effective method to fine-tune pre-trained text-to-image diffusion models on a small number of input images for personalized text-guided image generation. 
It stores the subject's appearance into the diffusion model weights with a specific text-token \textit{``[V]''}. 
Dreambooth is fine-tuned with two loss functions. 
Reconstruction loss is standard diffusion denoising supervision on the input images. 
Class-specific prior preservation loss is proposed to avoid language drift and loss of diversity caused by fine-tuning. 
It further supervises the pre-trained model with a large number of its own generated examples. 
TextureDreamer uses DreamBooth to distill texture information from input images. 
Instead of image synthesis, we apply the distilled information to a 3D object with different geometry.

\topic{ControlNet}~\cite{zhang2023adding} proposes a novel architecture that adds spatial conditioning control to pre-trained diffusion models. 
The key insight is to reuse the large number of diffusion model parameters trained on billions of images and insert small convolution networks into the model with window size 1 and zero-initialized weights. 
It enables robust fine-tuning performance on small datasets with different types of 2D conditions, such as depth, normal, and edge maps. 
We utilize ControlNet models to ensure that our created textures are aligned with the given geometry.

\topic{Score Distillation Sampling}~\cite{poole2022dreamfusion,wang2023score} is the core component of numerous methods that use pre-trained 2D diffusion models for 3D content creation \cite{poole2022dreamfusion,lin2023magic3d,chen2023fantasia3d}. 
It optimizes the 3D representation by pushing its rendered images to a high-dimensional manifold modeled by the pre-trained diffusion model. 
Let $\theta$ be the 3D representation and $\epsilon_{\psi}$ be the pre-trained diffusion model. 
The gradient back-propagated to the parameter $\theta$ is
\begin{equation}
    \nabla_{\theta}\mathcal{L}_{\text{SDS}}(\theta)
    \triangleq \mathbb{E}_{t,\epsilon} \left[w(t)(\epsilon_{\psi}\left(\mathbf{x}_t,y,t) - {\epsilon}\right) \frac{\partial g(\theta,c)}{\partial \theta} \right], \nonumber
\end{equation} 
where $\omega(t)$ is the weight coefficient, $y$ is the text input, $t$ is the time step, $c$ is the camera pose, $g(\cdot)$ is a differentiable renderer, $\mathbf{x}_t$ is the noisy image computed by adding noise to the rendered image $\mathbf{x}=g(\theta,c)$ with variance dependent on time $t$. 
Despite its wide usage, SDS requires a much higher weight than normal classifier-free guidance~\cite{ho2022classifier} to converge, oversmoothed and oversaturated appearance. 
To overcome this issue, Wang et al.~\cite{wang2023prolificdreamer} propose an improved version, called variational score distillation (VSD), which can converge with standard classifier-free guidance. 
VSD treats the whole 3D representation $\theta$ as a random variable and minimizes the KL divergence between $\theta$ and the distribution defined by the pre-trained diffusion model. It involves fine-tuning a LoRA~\cite{hu2021lora} network $\epsilon_\phi$ (and a camera encoder $\rho$ which embeds camera pose $c$ as an condition input to $\epsilon_\phi$) to denoise the noisy images generated from 3D representation $\theta$ 
\begin{equation}
\min_{\phi} \mathbb{E}_{t,\epsilon,c}\!\left[||\epsilon_\phi(\mathbf{x}_t,y,t,c)-\epsilon||_2^2\right] \label{eq:lora}
\end{equation}
The gradient to the 3D representation $\theta$ is then computed as
\begin{equation}
\mathbb{E}_{t\text,\epsilon,c} \left[w(t)({\epsilon}_{\psi}\left(\mathbf{x}_t,y,t) - {\epsilon}_{\phi}(\mathbf{x}_t, y,t,c)\right) \frac{\partial g(\theta, c)}{\partial \theta}\right]. \label{eq:vsd_gradient}
\end{equation}
While VSD significantly improves both visual quality and diversity of generated 3D contents, it cannot address the 3D consistency issue due to the inherent lack of 3D knowledge, leading to multi-face errors and mismatches between geometry and textures. 
We address this challenge by explicitly injecting geometry information to make our diffusion model geometry aware.

\subsection{Personalized Geometry-aware Score Distillation (PGSD)}\label{sec:pgsd} 

\topic{Problem setup.} 
We illustrate our method in Figure \ref{fig:method}. 
The inputs to our framework include a small set of images (3 to 5) casually captured from different views $\{I\}_{k=1}^{K}$ and a target 3D mesh $\mathcal{M}$. 
The outputs of our framework are relightable textures transferred from image set $\{I\}_{k=1}^{K}$ to $\mathcal{M}$ in a semantically meaningful and visually pleasing manner. 
Our relightable textures are parameterized as standard microfacet bidirectional reflectance distribution (BRDF) model \cite{karis2013real}, which consists of 3 parameters, diffuse albedo $a$, roughness $r$, and metallic $m$. 
We deliberately \emph{do not} optimize normal maps as it encourages the pipeline to fake details that are inconsistent with mesh $\mathcal{M}$. 
Following the recent trend of neural implicit representation \cite{muller2022instant,munkberg2022extracting,hasselgren2022shape}, during optimization, we represent our texture as a neural BRDF field $f_{\theta}(v): v \in \mathbb{R}^3 \rightarrow, a, r, m \in \mathbb{R}^5$, where $v$ is an arbitrary point sampled on the surface of $\mathcal{M}$ and $f_{\theta}$ consists of a multi-scale hash encoding and a small MLP. 
We find such an implicit representation can better regularize the optimization process, leading to smoother textures. 
However, given the UV mapping of $\mathcal{M}$, our representation can also be converted to standard 2D texture maps that are compatible with standard graphics pipelines, by querying every 3D point corresponding to each texel, as shown on the right-hand side of Figure \ref{fig:method}.

\topic{Personalized texture information extraction.} 
We follow Dreambooth \cite{ruiz2023dreambooth} to extract texture information from sparse images.
To be specific, we fine-tune a personalized diffusion model on input images with a text prompt $y$, \textit{``A photo of [V] object''}, where \textit{``[V]''} is a unique identifier to describe the input object.
Compared to the alternative textual inversion method \cite{gal2022image}, we observe that Dreambooth converges faster and can better preserve intricate texture patterns, possibly due to its larger capacity. 
We first mask out the background of the target object with a white color. 
For the reconstruction loss,  we resize the shorter edge of input images to 512 and randomly crop 512x512 patches for training. 
We do not apply class-specific prior preservation loss, as we hope our Dreambooth finetuning model can generalize to other categories.
We also experiment with different variations, including jointly fine-tuning the text encoder and replacing the diffusion denoising network with a pre-trained ControlNet, but do not observe any improvements. 

\topic{Geometry-aware score distillation} 
Once we finish extracting texture information with Dreambooth, we transfer the information to mesh $\mathcal{M}$ by adopting the fine-tuned Dreambooth model as the denoising network $\epsilon_{\psi}$ for score distillation sampling. 
Specifically, we choose VSD instead of the original SDS because of its superior ability to generate highly realistic and diverse appearances. 
To render images $\mathbf{x}$ for VSD gradient computation, we follow Fantasia3D~\cite{chen2023fantasia3d} to pre-select a fixed HDR environment map $E$ as illumination and use Nvdiffrast \cite{Laine2020diffrast} as our differentiable renderer. 
We set the object background to be a constant white color to match the input images for Dreambooth training. 
We observe this can help achieve better color fidelity compared to random color or neutral background. 

However, simply replacing SDS with VSD cannot address the limitation of lacking 3D knowledge in 2D diffusion models. 
We thus propose geometry-aware score distillation, where we inject geometry information extracted from mesh $\mathcal{M}$ into our personalized diffusion model $\epsilon_\psi$ through a pre-trained ControlNet conditioned on normal maps $k$ rendered from $\mathcal{M}$. 
This augmentation significantly boosts 3D consistency of generated textures (see Figure~\ref{fig:ablation}). 
With the ControlNet conditioning, the pillow texture from the input images can be accurately matched to the target shape, despite the shape mismatch. 
We experiment with different ControlNet conditions and show that normal conditions can best prevent texture-geometry mismatch.

Let $\textbf{x}=g(\theta,c)$ be the rendered image under a fixed environment map from camera pose $c$ with the extracted BRDF maps $a_{\theta}, r_{\theta}, m_{\theta}$. 
The gradient of proposed Personalized Geometry-aware Score Distillation (PGSD) to optimize the MLP parameter $\theta$ of BRDF field is:
\begin{equation}
\begin{aligned}
       \nabla_{\theta}&\mathcal{L}_{\text{PGSD}}(\theta) \\
   &\triangleq \mathbb{E}_{t,\epsilon,c} [w(t)({\epsilon}_\psi(\mathbf{x}_t;y,k,t) -{\epsilon}_\phi(\mathbf{x}_t;y,k,t,c_{\rho}) ) \frac{\partial \textbf{x}}{\partial \theta} ], \nonumber
\end{aligned}
\end{equation}
where $\textbf{x}_t=\alpha_{t}\mathbf{x}+\sigma_t \epsilon$ is the rendered image $\textbf{x}$ perturbed by noise $\epsilon\sim \mathcal{N}(\mathbf{0}, \mathbf{I})$ at time $t$, $c_{\rho}$ is the embedding of the camera extrinsic $c$ encoded by a learnable camera encoder $\rho$, $\epsilon_{\psi}$ and $\epsilon_{\phi}$ are the fine-tuned personalized diffusion model and the generic diffusion model pretrained on a large-scale dataset, respectively. Both models are augmented with ControlNet conditioned on normal map $k$, as shown in the yellow part underneath the diffusion model in Figure~\ref{fig:method}.

We found that our method does not benefit from classifier-free guidance (CFG)~\cite{ho2022classifier}, probably because the personalized model $\epsilon_{\psi}$ has been fine-tuned on a small number of images. Since our goal is to faithfully transfer input appearance to target shape, it is not necessary to have CFG to increase the diversity. Similar observation can be found in recent literature~\cite{sharma2023alchemist}.

We additionally identify several important design choices through extensive experiments. 
First, it is important to initialize the $\epsilon_{\phi}$ in Eq. \ref{eq:lora} with original pre-trained diffusion model weights while the Dreambooth weight will remove texture details. 
This is probably because the Dreambooth fine-tuning process makes the diffusion model overfit to a small training set, as pointed out by previous work \cite{raj2023dreambooth3d}. 
Moreover, we find that removing the LoRA weights can substantially improve texture fidelity. 
Similar difficulties in training LoRA were also reported in \cite{shi2023zero123plus}. 
We therefore implement our personalized geometry-aware score distillation loss $\mathcal{L}_{PGSD}$ by removing the LoRA structure in $\epsilon_{\phi}$ and only keeping the camera embedding, achieving the best quality. 
We show more comparisons in Figure \ref{fig:ablation}.

\section{Experiment}

\subsection{Experimental setup}
\topic{Dataset.}
We conduct our experiments on 4 categories of objects: sofa, bed, mug/bowl, and plush toy. For each category, we select 8 instances of objects and create a small image set by casually sampling 3 to 5 views surrounding the object, resulting in 32 image sets in total.
For every image in the 32 image sets, we apply U2-Net~\cite{Qin_2020_PR} to obtain the foreground mask automatically or use a semi-auto background removal application\footnote{https://www.remove.bg/upload} to obtain more accurate masks. 
We perform texture transfer for each image set to diverse meshes including but not limited to same category shapes, different category shapes, or even geometry with different genus numbers.  To test our texture-transferring framework, we select 3 meshes for each of the 4 categories that are dissimilar to the captured image sets. We acquire these 3D meshes from 3D-FUTURE~\cite{fu20213d} and online repositories.\footnote{https://www.cgtrader.com/}\footnote{https://sketchfab.com/}.  We run intra-class texture transfer for all 4 categories of objects and also run inter-class texture transfer between bed and chair, to test our method's generalization ability.

\begin{figure*}[t]
  \centering
  \includegraphics[width=1.0\linewidth]{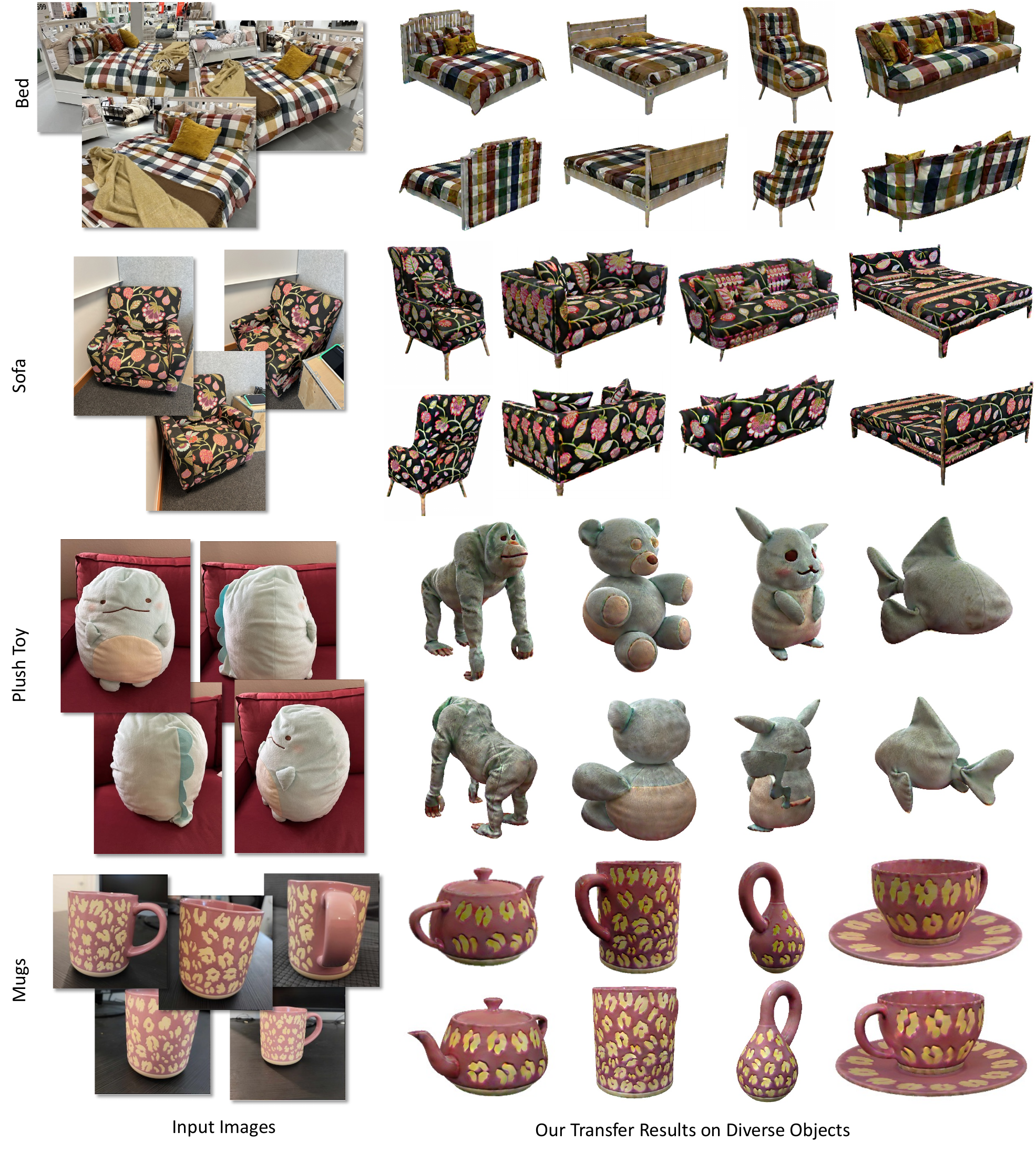}
\vspace{-3mm}
  \caption{\textbf{Image-guided transfer results} from four categories (beds, sofas, plush toys, and mugs) of image sets to diverse objects. Our method can be applied to a wide range of object types and transfer the textures to diverse object shapes. }
  
  \label{fig:results}
\end{figure*}

\topic{Implementation details.} 
We implement our framework based on PyTorch~\cite{paszke2019pytorch} and Threestudio~\cite{threestudio2023}. 
We use latent diffusion and ControlNet v1.1 as our pre-trained diffusion model and ControlNet respectively. 
In all our experiments, we set the classifier-free guidance weight of $\mathcal{L}_{PGSD}$ as 1.0 (equivalent to setting $\omega=0$ in the original CFG formulation).
Following DreamFusion~\cite{poole2022dreamfusion}, we also apply view-dependent conditioning to the input text prompt.
The BRDF field is parameterized with an MLP using hash-grid positional encoding~\cite{muller2022instant}, following prior works~\cite{chen2023fantasia3d,wang2023prolificdreamer}. 
Our camera encoder consists of two linear layers that project the camera extrinsic to a latent vector of $1,280$ dimensions to be fused with time and text embedding in U-Net. 
We empirically set the learning rate to $0.01$ for encoding, $0.001$ for MLP, and $0.0001$ for camera encoder for all experiments.

\subsection{Baseline methods}
Latent-paint~\cite{metzer2023latent} and TEXTure~\cite{richardson2023texture} are two recent text-guided texturing methods with 2D diffusion prior.
They also demonstrate the capability of texturing meshes from images. 
Latent-paint~\cite{metzer2023latent} leverages the Texture Inversion~\cite{gal2022image} to extract image information into text embedding and distills the texture with SDS.
TEXTure~\cite{richardson2023texture} first finetunes the pre-trained diffusion model by combining Texture Inversion and Dreambooth~\cite{ruiz2023dreambooth} and use this fine-tuned model to synthesize texture with an iterative mesh painting algorithm. As preferred by the previous method~\cite{richardson2023texture}, we augment the input images with a random color background. We closely follow the original implementation of baseline methods to run the experiments.

\begin{figure*}[t]
  \centering
  \includegraphics[width=0.95\linewidth]{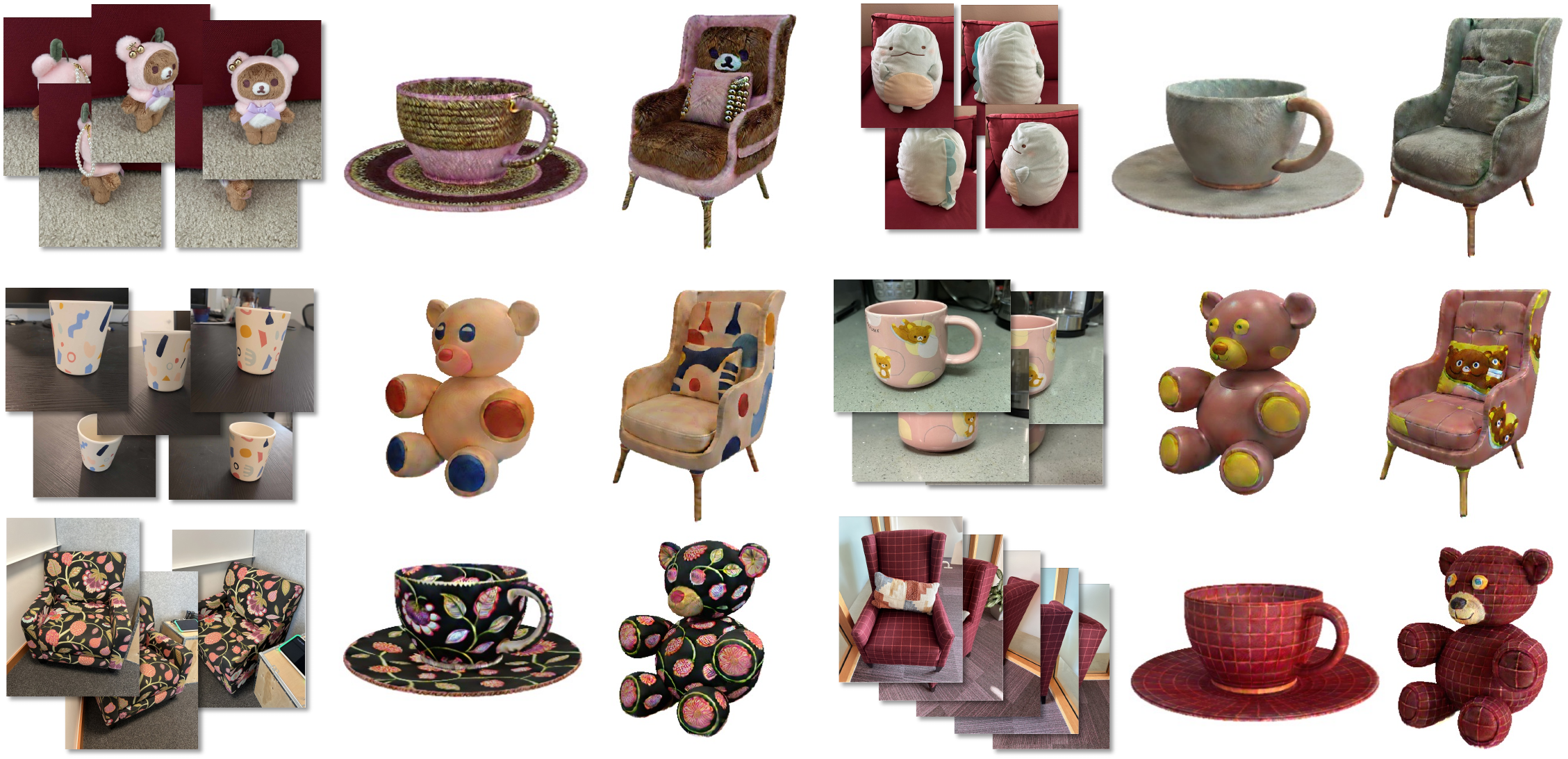}

  \caption{\textbf{Example of cross-category texture transfer results.} In the first row, we transfer appearances from plush toys to cups and chairs. In the second row, special patterns from mugs are transferred to bears and chairs. In the thrid row, textures from input sofa are transferred to cups and bears.}
  
  \label{fig:exp_cross}
\end{figure*}
\begin{figure}[t]
  \centering
  \includegraphics[width=1.0\linewidth]{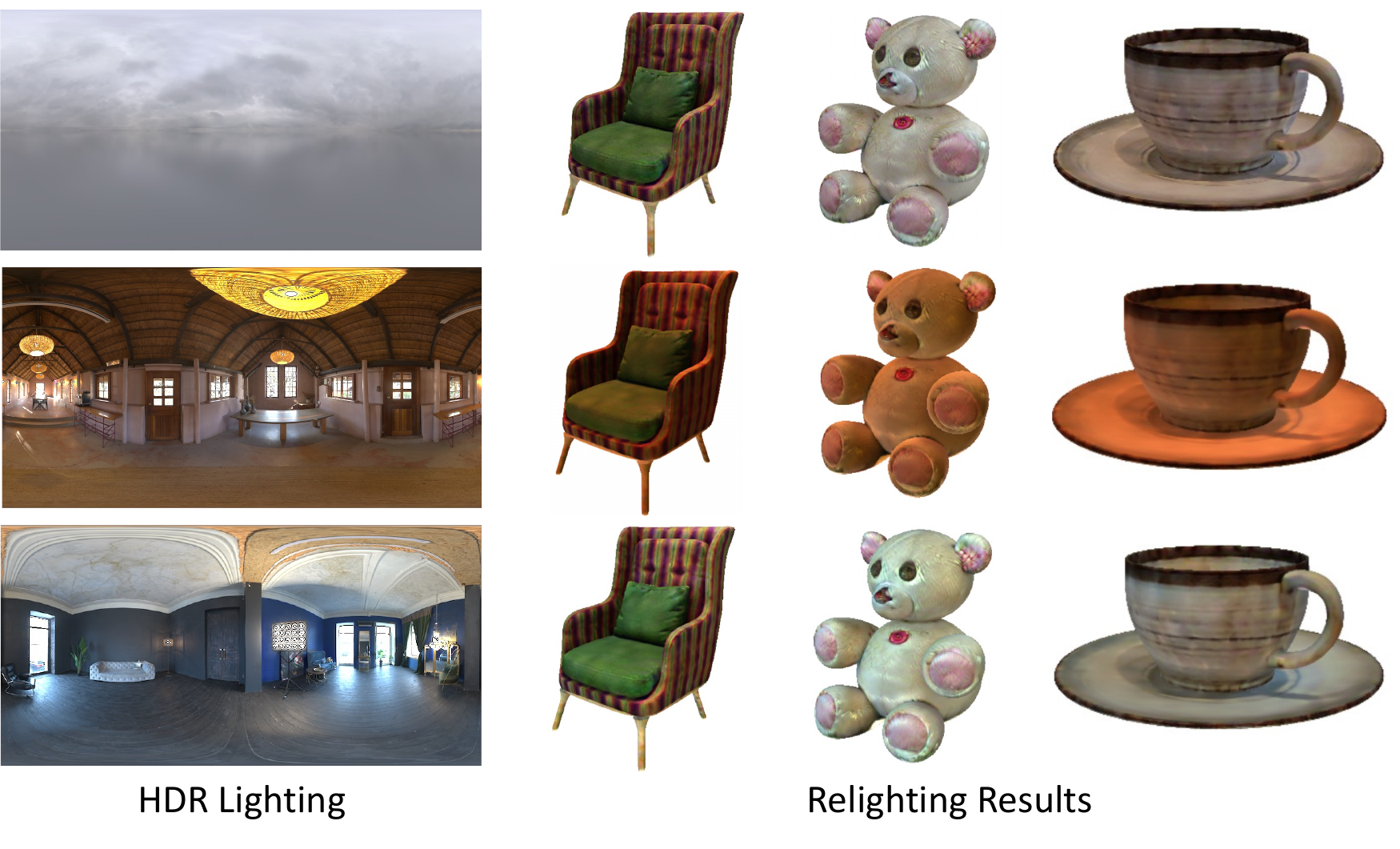}

  \caption{\textbf{Example of relighting results.} The textures are relit by the original HDR environment maps (first row) and the novel maps (second and third rows). }
  
  \label{fig:exp_relight}
\end{figure}

\begin{figure*}[t]
  \centering
  \includegraphics[width=0.9\linewidth]{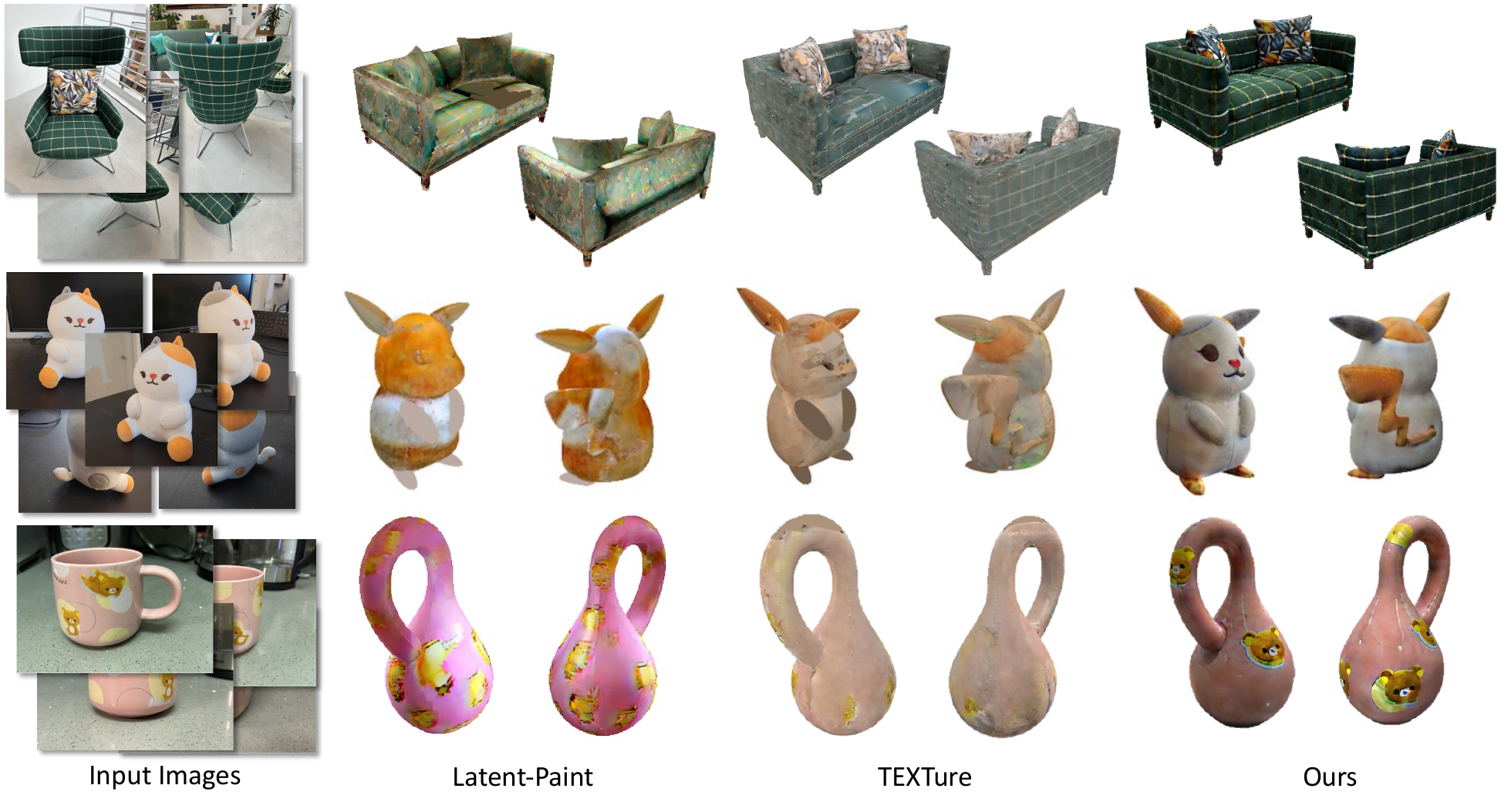}
  \vspace{-3mm}
  \caption{\textbf{Comparison between baseline methods. }Compared with Latent-Paint~\cite{metzer2023latent} and TEXTure~\cite{richardson2023texture}, our method can synthesize seamless and geometry-aware textures which are compatible with the target mesh geometry.}
  
  \label{fig:results_baseline}
\end{figure*}
\begin{figure}[t]
  \centering
  \includegraphics[width=1.0\linewidth]{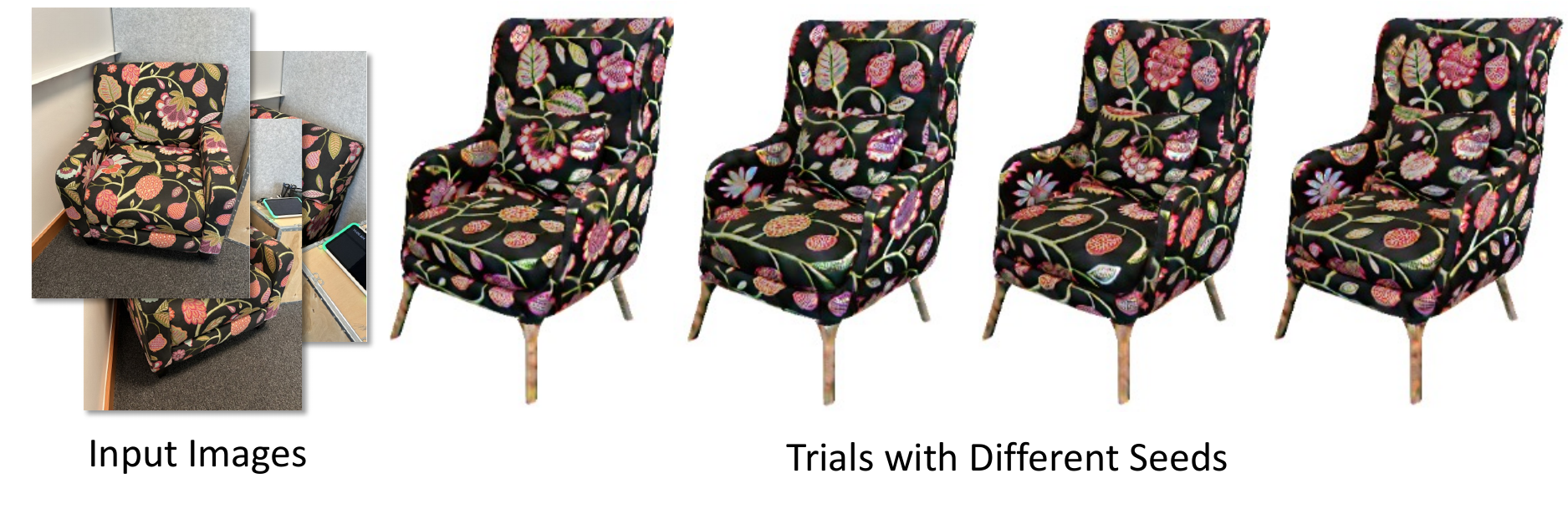}

  \caption{\textbf{Diversity} of synthesized textures.}
  \label{fig:seed_diversity}
\end{figure}
\begin{figure}[t]
  \centering
  \includegraphics[width=1.0\linewidth]{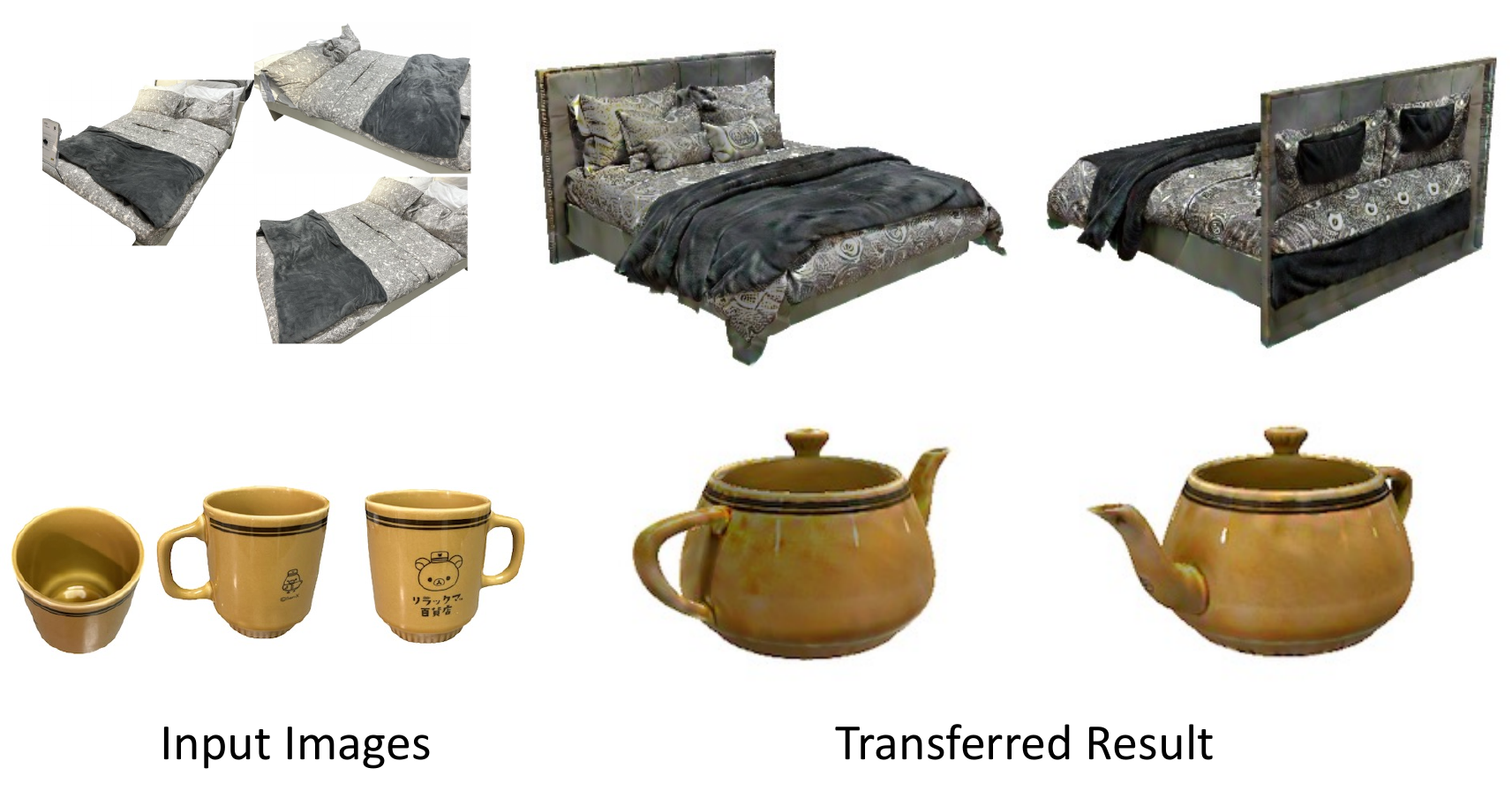}
  \caption{\textbf{Limitations.} Our method may bake-in lighting into texture, have Janus problem when lacking enough input viewpoints, and ignore special and non-repeated patterns from the input.}
  \label{fig:failure}
\end{figure}
\subsection{Image-guided texture transfer}
\label{sec:exp_image_guided}
\begin{table}[]
\centering
\begin{tabular}{lcc}
\cline{2-3}
                          & \multicolumn{2}{c}{Ours preferred over} \\ \cline{2-3} 
                          & Latent-Paint          & TEXTure         \\ \hline
Image Fidelity            & 71.82\%                & 69.43\%          \\
Texture Photorealism      & 77.03\%                & 85.52\%          \\
Shape-Texture Consistency & 78.49\%                & 85.16\%          \\ \hline
\end{tabular}
\caption{\textbf{User study} on image-guided texture transfer.}
\label{tab:image_guided_user}
\end{table}

\begin{table}[]
    \centering
    \begin{tabular}{lc}
    \hline
                 & CLIP similarity $\uparrow$ \\ \hline
    Latent-Paint~\cite{metzer2023latent} & 0.7969                 \\
    TEXTure~\cite{richardson2023texture}      & 0.7988                 \\
    Ours         & \textbf{0.8296}                 \\ \hline
    \end{tabular}
    \caption{\textbf{Quantitative evaluation} on image-guided texturing.}
    \label{tab:image_guided_quan}
\end{table}
\vspace{-3mm}

\topic{Qualitative evaluation}
Our method can perform texture transfer to diverse object geometry, including geometry in the \emph{same} category or across {different} categories. 
Figure~\ref{fig:results} demonstrates our texture transferring results on 4 categories of objects.
Our method can synthesize geometry-aware and seamless textures that has similar patterns and styles as the input.
We also demonstrate that our method can transfer textures \emph{across different categories}. 
In Figure~\ref{fig:teaser}, we show texture transfer results from images of sofa to bed shapes, and vice versa.
Our method is also capable of performing texture transfers across a broader range of different categories. As shown in Figure~\ref{fig:exp_cross}, high-quality and realistic textures can be synthesized across chair, mug, and plush toy categories.
Since our synthesized texture contains albedo, metallic, and roughness maps,
the target objects with the synthesized appearance can be relit, as shown in Figure~\ref{fig:exp_relight}.
By using different random seeds, our framework can generate diverse textures, as shown in Figure~\ref{fig:seed_diversity}.

In Figure~\ref{fig:results_baseline}, we qualitatively compare our method with baseline methods. 
Two views are shown in each example.
Latent-Paint tends to generate textures with colors and patterns that are different from input images. 
TEXTure can preserve the color and texture better than Latent-Paint, but the texture contains visible seams (possibly due to the iterative painting). 
Our results method can reason the semantics of the geometry (\eg the positions of eyes) and demonstrate higher quality, seamless, and geometry-aware texturing results with higher fidelity from the input images. 

\topic{Quantitative evaluation}
It is non-trivial to perform quantitative comparisons for texture transfer due to the shape difference between geometry and photos.
We perform a user study to evaluate transfer fidelity, texture photorealism, and texture-geometry compatibility across baselines by asking users the following questions: 
1) Which one has the texture that looks more similar to input images? 
2) Which one has a texture which looks more like a real object? 
3) Which one has the texture which is more compatible with the meshes? (Which texture painted more fitted to the geometry?) 
We conduct a user study with Amazon Turk with three separate tasks. 
For each task, we ask each user 24 questions. 
Each question is a forced single-choice selection with two options among our and one baseline result with the rendered images from the same $4$ sampled views and is evaluated by 20 different users. 
We only show input photos for the first similarity question, and hide the input photos for the other two questions to make the user focus on texture quality. 
We summarize the results in Table~\ref{tab:image_guided_user}.
Our results show significant preference by the users in terms of image fidelity, texture photorealism, and shape-texture consistency.

We also propose to evaluate the similarity via image-based CLIP feature~\cite{mohammad2022clip}
between reference and the rendered images of synthesized textures. 
The CLIP similarity has been applied to material matching~\cite{Yan:2023:PSDR-Room} and stylization~\cite{text2mesh}. 
A good transfer should transfer only the texture from images and should take into account the target shape geometry and transfer the texture semantically. 
For example, the transfer should be painted with respect to each part of the shape. 
We use our evaluation set to compute the comparison. 
For each image set and target 3D mesh pair, we compute the average of the metric among each reference image and each of rendered image from $4$ sampled views (\ie left front, right front, left back, and right back). 
We average the CLIP similarity across all (image set, mesh) pairs.
Table~\ref{tab:image_guided_quan} shows our method has the highest CLIP similarity.

\begin{figure*}[t]
  \centering
  \includegraphics[width=1.0\linewidth]{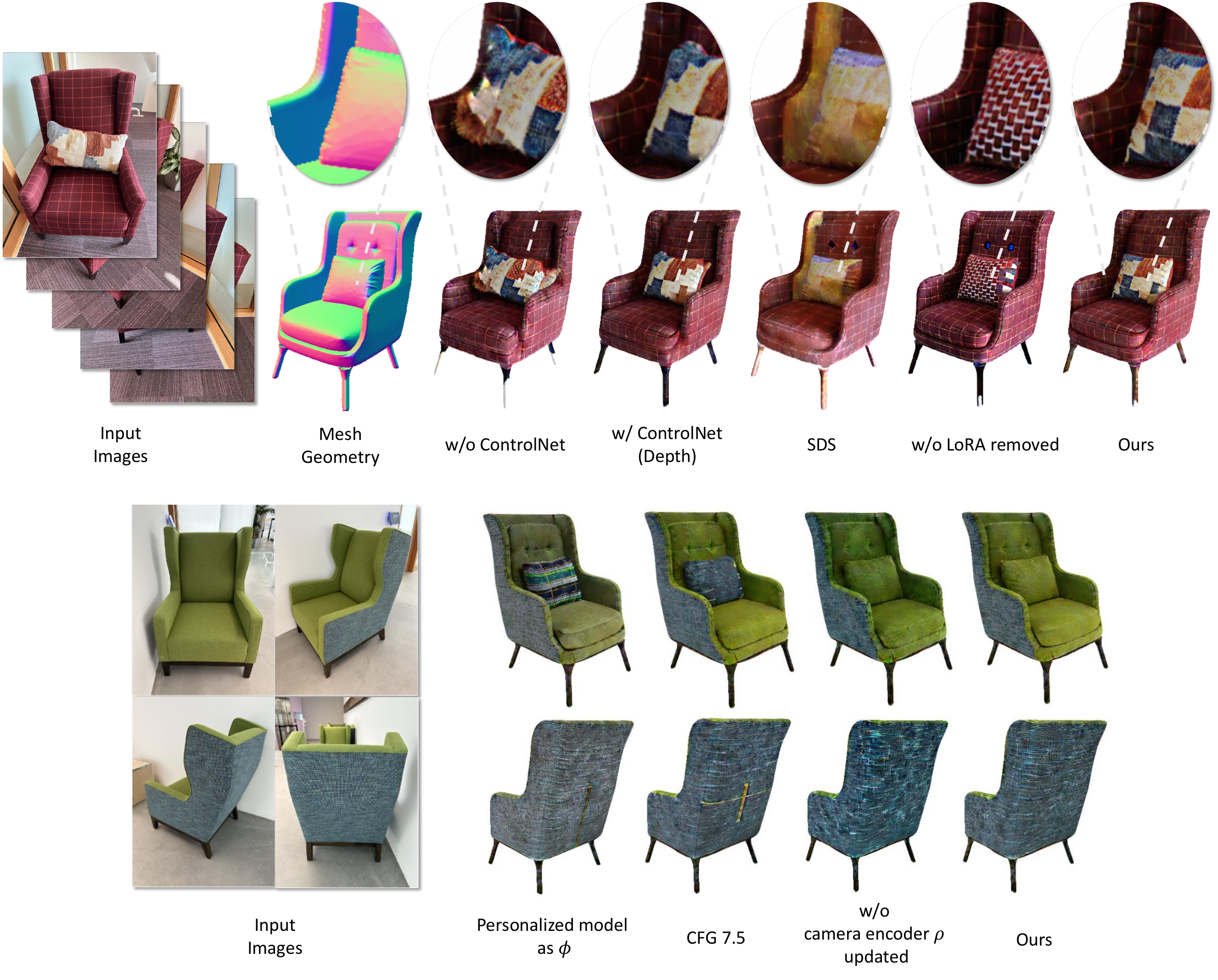}
  \vspace{-3mm}
  \caption{\textbf{Ablation study.} 
  (First row) With ControlNet conditioned on normal maps, the result has the best texture-geometry consistency. Without ControlNet or with depth-based ControlNet, the results suffer from texture-geometry misalignment. 
  Using SDS loss leads to blurry textures. 
  Without the LoRA module removed, the results tend to remove the existing texture from the personalized diffusion model. Our full method can synthesize accurate texture which is similar to input appearances. (Second row) If replacing generic diffusion model $\phi$ with personalized model or applying classifier guidance scale $7.5$, some random patterns might appear in the synthesized texture. If we freeze the camera encoder $\rho$, the result might be worse or more noisy than our full method.}
  \label{fig:ablation}
\end{figure*}

\subsection{Ablation Studies}
\label{sec:ablation}

We first qualitatively perform an ablation study on the importance of geometry-aware ControlNet. 
As shown in Figure~\ref{fig:ablation}, the results suffer from geometry-texture misalignment without ControlNet or the depth-based ControlNet. 
Only normal-based ControlNet can accurately control the synthesized texture to be consistent with the input mesh geometry.
Next, we validate the importance of score distillation loss. 
Only using SDS loss in our framework cannot achieve enough input fidelity and the result tends to be more blurry. 
Without LoRA removed (which is usually optimized with vanilla VSD loss), the optimization tends to make the distribution diverge from the Dreambooth-finetuned distribution. 
This results in the output containing less original texture but more irrelevant patterns from the input. 
We hypothesize that this is due to optimizing LoRA weights with a text condition containing a rare identifier tends to drive the distribution of rendered images to have a rare appearance.

If we replace generic diffusion model $\epsilon_\phi$ with the personalized diffusion model $\epsilon_\psi$ or apply classifier free guidance weight $7.5$, the result tends to introduce random patterns which does not exist in the input images. If we choose to freeze the camera encoder weights $\rho$, the result becomes worse or more noisy than our full method.

We also quantitatively evaluate the importance of each component in our system. We use image-based CLIP feature to measure the similarity between reference images and the rendered images. To ensure fair evaluation, the background of both reference and rendered images are masked with white color.
\begin{table}[]
    \centering
    \caption{\textbf{Ablation study} on image-based texturing w.r.t. CLIP image-based feature similarity. Although \textit{w/o ControlNet} and \textit{w/ ControlNet (Depth)} achieve higher similarity score, the transfer results tend to ignore target shape and directly paint the texture without reasoning the geometry. Among the remaining ablative methods, our full method achieves the highest CLIP similarity w.r.t. reference images. }
    \begin{tabular}{lc}
    \hline
                   & CLIP similarity $\uparrow$ \\ \hline
    w/o ControlNet               & \textit{0.8394}                 \\
    w/ ControlNet (Depth)        & \textit{0.8320}                \\
    SDS, w/o CFG                 & 0.8101                \\
    SDS, CFG $100$               & 0.7983                \\
    w/o LoRA removed             & 0.8110                 \\
    Personalized model as $\phi$ & 0.8218                 \\
    CFG weight as $7.5$          & 0.8218                 \\
    w/o camera encoder $\rho$ updated & 0.8267                 \\
    Ours                         & \textbf{0.8296}                 \\ \hline
    \end{tabular}
    
    \label{tab:ablation}
\end{table}
As shown in Table~\ref{tab:ablation}, our full method achieves the highest similarity score among the ablative baselines except \textit{w/o ControlNet} and \textit{w/ ControlNet (Depth)}. As shown in Figure~\ref{fig:ablation}, these two methods tend to ignore the target shape and directly paint the texture without adapting to geometry. Thus, they could reach higher score by painting the original texture regardless of the shape. We also observe that SDS results tend to be saturated or blurry and cannot recover the texture from the inputs. Keeping LoRA in the generic diffusion model $\epsilon_\phi$ will introduce random patterns to the synthesized texture.  

\section{Discussions}
We proposed a framework to transfer texture from input images to an arbitrary shape. 
While our method can transfer high-quality texture in most cases, there are some limitations.  
Figure~\ref{fig:failure} shows that our method may not be able to transfer special and non-repeated texture to the target shapes.
In addition, our method tends to bake in lighting to texture when there are strong specular highlights in the input images. 
Janus problem might appear when the viewpoints of input images do not cover the entire object.
Nevertheless, we believe that our method can be the first step to tackling this challenging problem and will make an impact in the 3D content creation community.

\clearpage
{
    \small
    \bibliographystyle{ieeenat_fullname}
    \bibliography{main}

\begin{thebibliography}{68}
\providecommand{\natexlab}[1]{#1}
\providecommand{\url}[1]{\texttt{#1}}
\expandafter\ifx\csname urlstyle\endcsname\relax
  \providecommand{\doi}[1]{doi: #1}\else
  \providecommand{\doi}{doi: \begingroup \urlstyle{rm}\Url}\fi

\bibitem[sub()]{substance}
{Adobe} substance 3d.
\newblock \url{https://docs.substance3d.com/sat.}

\bibitem[AlBahar et~al.(2023)AlBahar, Saito, Tseng, Kim, Kopf, and Huang]{albahar2023single}
Badour AlBahar, Shunsuke Saito, Hung-Yu Tseng, Changil Kim, Johannes Kopf, and Jia-Bin Huang.
\newblock Single-image 3d human digitization with shape-guided diffusion.
\newblock In \emph{SIGGRAPH Asia}, 2023.

\bibitem[Bi et~al.(2017)Bi, Kalantari, and Ramamoorthi]{bi2017patch}
Sai Bi, Nima~Khademi Kalantari, and Ravi Ramamoorthi.
\newblock Patch-based optimization for image-based texture mapping.
\newblock \emph{ACM Trans. Graph.}, 36\penalty0 (4):\penalty0 106--1, 2017.

\bibitem[Bokhovkin et~al.(2023)Bokhovkin, Tulsiani, and Dai]{bokhovkin2023mesh2tex}
Alexey Bokhovkin, Shubham Tulsiani, and Angela Dai.
\newblock Mesh2tex: Generating mesh textures from image queries.
\newblock \emph{arXiv preprint arXiv:2304.05868}, 2023.

\bibitem[Cai et~al.(2022)Cai, Yan, Dong, Gkioulekas, and Zhao]{cai2022physics}
G. Cai, K. Yan, Z. Dong, I. Gkioulekas, and S. Zhao.
\newblock Physics-based inverse rendering using combined implicit and explicit geometries.
\newblock \emph{Computer Graphics Forum}, 41\penalty0 (4):\penalty0 129--138, 2022.

\bibitem[Cao et~al.(2023)Cao, Kreis, Fidler, Sharp, and Yin]{cao2023texfusion}
Tianshi Cao, Karsten Kreis, Sanja Fidler, Nicholas Sharp, and Kangxue Yin.
\newblock Texfusion: Synthesizing 3d textures with text-guided image diffusion models.
\newblock In \emph{Proceedings of the IEEE/CVF International Conference on Computer Vision}, pages 4169--4181, 2023.

\bibitem[Chan et~al.(2022)Chan, Lin, Chan, Nagano, Pan, De~Mello, Gallo, Guibas, Tremblay, Khamis, et~al.]{chan2022efficient}
Eric~R Chan, Connor~Z Lin, Matthew~A Chan, Koki Nagano, Boxiao Pan, Shalini De~Mello, Orazio Gallo, Leonidas~J Guibas, Jonathan Tremblay, Sameh Khamis, et~al.
\newblock Efficient geometry-aware 3d generative adversarial networks.
\newblock In \emph{Proceedings of the IEEE/CVF Conference on Computer Vision and Pattern Recognition}, pages 16123--16133, 2022.

\bibitem[Chang et~al.(2015)Chang, Funkhouser, Guibas, Hanrahan, Huang, Li, Savarese, Savva, Song, Su, et~al.]{chang2015shapenet}
Angel~X Chang, Thomas Funkhouser, Leonidas Guibas, Pat Hanrahan, Qixing Huang, Zimo Li, Silvio Savarese, Manolis Savva, Shuran Song, Hao Su, et~al.
\newblock Shapenet: An information-rich 3d model repository.
\newblock \emph{arXiv preprint arXiv:1512.03012}, 2015.

\bibitem[Chen et~al.(2023{\natexlab{a}})Chen, Siddiqui, Lee, Tulyakov, and Nie{\ss}ner]{chen2023text2tex}
Dave~Zhenyu Chen, Yawar Siddiqui, Hsin-Ying Lee, Sergey Tulyakov, and Matthias Nie{\ss}ner.
\newblock Text2tex: Text-driven texture synthesis via diffusion models.
\newblock \emph{arXiv preprint arXiv:2303.11396}, 2023{\natexlab{a}}.

\bibitem[Chen et~al.(2023{\natexlab{b}})Chen, Chen, Jiao, and Jia]{chen2023fantasia3d}
Rui Chen, Yongwei Chen, Ningxin Jiao, and Kui Jia.
\newblock Fantasia3d: Disentangling geometry and appearance for high-quality text-to-3d content creation.
\newblock \emph{arXiv preprint arXiv:2303.13873}, 2023{\natexlab{b}}.

\bibitem[Chen et~al.(2022)Chen, Yin, and Fidler]{chen2022auv}
Zhiqin Chen, Kangxue Yin, and Sanja Fidler.
\newblock Auv-net: Learning aligned uv maps for texture transfer and synthesis.
\newblock In \emph{Proceedings of the IEEE/CVF Conference on Computer Vision and Pattern Recognition}, pages 1465--1474, 2022.

\bibitem[Choy et~al.(2016)Choy, Xu, Gwak, Chen, and Savarese]{choy20163d}
Christopher~B Choy, Danfei Xu, JunYoung Gwak, Kevin Chen, and Silvio Savarese.
\newblock 3d-r2n2: A unified approach for single and multi-view 3d object reconstruction.
\newblock In \emph{Computer Vision--ECCV 2016: 14th European Conference, Amsterdam, The Netherlands, October 11-14, 2016, Proceedings, Part VIII 14}, pages 628--644. Springer, 2016.

\bibitem[Efros and Leung(1999)]{efros1999texture}
Alexei~A Efros and Thomas~K Leung.
\newblock Texture synthesis by non-parametric sampling.
\newblock In \emph{Proceedings of the seventh IEEE international conference on computer vision}, pages 1033--1038. IEEE, 1999.

\bibitem[Erko{\c{c}} et~al.(2023)Erko{\c{c}}, Ma, Shan, Nie{\ss}ner, and Dai]{erkocc2023hyperdiffusion}
Ziya Erko{\c{c}}, Fangchang Ma, Qi Shan, Matthias Nie{\ss}ner, and Angela Dai.
\newblock Hyperdiffusion: Generating implicit neural fields with weight-space diffusion.
\newblock \emph{arXiv preprint arXiv:2303.17015}, 2023.

\bibitem[Fu et~al.(2021)Fu, Jia, Gao, Gong, Zhao, Maybank, and Tao]{fu20213d}
Huan Fu, Rongfei Jia, Lin Gao, Mingming Gong, Binqiang Zhao, Steve Maybank, and Dacheng Tao.
\newblock 3d-future: 3d furniture shape with texture.
\newblock \emph{International Journal of Computer Vision}, 129:\penalty0 3313--3337, 2021.

\bibitem[Gal et~al.(2022)Gal, Alaluf, Atzmon, Patashnik, Bermano, Chechik, and Cohen-or]{gal2022image}
Rinon Gal, Yuval Alaluf, Yuval Atzmon, Or Patashnik, Amit~Haim Bermano, Gal Chechik, and Daniel Cohen-or.
\newblock An image is worth one word: Personalizing text-to-image generation using textual inversion.
\newblock In \emph{The Eleventh International Conference on Learning Representations}, 2022.

\bibitem[Gao et~al.(2022)Gao, Shen, Wang, Chen, Yin, Li, Litany, Gojcic, and Fidler]{gao2022get3d}
Jun Gao, Tianchang Shen, Zian Wang, Wenzheng Chen, Kangxue Yin, Daiqing Li, Or Litany, Zan Gojcic, and Sanja Fidler.
\newblock Get3d: A generative model of high quality 3d textured shapes learned from images.
\newblock \emph{Advances In Neural Information Processing Systems}, 35:\penalty0 31841--31854, 2022.

\bibitem[Goodfellow et~al.(2020)Goodfellow, Pouget-Abadie, Mirza, Xu, Warde-Farley, Ozair, Courville, and Bengio]{goodfellow2020generative}
Ian Goodfellow, Jean Pouget-Abadie, Mehdi Mirza, Bing Xu, David Warde-Farley, Sherjil Ozair, Aaron Courville, and Yoshua Bengio.
\newblock Generative adversarial networks.
\newblock \emph{Communications of the ACM}, 63\penalty0 (11):\penalty0 139--144, 2020.

\bibitem[Guo et~al.(2023)Guo, Liu, Shao, Laforte, Voleti, Luo, Chen, Zou, Wang, Cao, and Zhang]{threestudio2023}
Yuan-Chen Guo, Ying-Tian Liu, Ruizhi Shao, Christian Laforte, Vikram Voleti, Guan Luo, Chia-Hao Chen, Zi-Xin Zou, Chen Wang, Yan-Pei Cao, and Song-Hai Zhang.
\newblock threestudio: A unified framework for 3d content generation.
\newblock \url{https://github.com/threestudio-project/threestudio}, 2023.

\bibitem[Hasselgren et~al.(2022)Hasselgren, Hofmann, and Munkberg]{hasselgren2022shape}
Jon Hasselgren, Nikolai Hofmann, and Jacob Munkberg.
\newblock Shape, light, and material decomposition from images using monte carlo rendering and denoising.
\newblock \emph{Advances in Neural Information Processing Systems}, 35:\penalty0 22856--22869, 2022.

\bibitem[Henderson et~al.(2020)Henderson, Tsiminaki, and Lampert]{henderson2020leveraging}
Paul Henderson, Vagia Tsiminaki, and Christoph~H Lampert.
\newblock Leveraging 2d data to learn textured 3d mesh generation.
\newblock In \emph{Proceedings of the IEEE/CVF conference on computer vision and pattern recognition}, pages 7498--7507, 2020.

\bibitem[Ho and Salimans(2022)]{ho2022classifier}
Jonathan Ho and Tim Salimans.
\newblock Classifier-free diffusion guidance.
\newblock \emph{arXiv preprint arXiv:2207.12598}, 2022.

\bibitem[Ho et~al.(2020)Ho, Jain, and Abbeel]{ho2020denoising}
Jonathan Ho, Ajay Jain, and Pieter Abbeel.
\newblock Denoising diffusion probabilistic models.
\newblock \emph{Advances in neural information processing systems}, 33:\penalty0 6840--6851, 2020.

\bibitem[Hu et~al.(2021)Hu, Shen, Wallis, Allen-Zhu, Li, Wang, Wang, and Chen]{hu2021lora}
Edward~J Hu, Yelong Shen, Phillip Wallis, Zeyuan Allen-Zhu, Yuanzhi Li, Shean Wang, Lu Wang, and Weizhu Chen.
\newblock Lora: Low-rank adaptation of large language models.
\newblock \emph{arXiv preprint arXiv:2106.09685}, 2021.

\bibitem[Karis and Games(2013)]{karis2013real}
Brian Karis and Epic Games.
\newblock Real shading in unreal engine 4.
\newblock \emph{Proc. Physically Based Shading Theory Practice}, 4\penalty0 (3):\penalty0 1, 2013.

\bibitem[Karnewar et~al.(2023)Karnewar, Vedaldi, Novotny, and Mitra]{karnewar2023holodiffusion}
Animesh Karnewar, Andrea Vedaldi, David Novotny, and Niloy~J Mitra.
\newblock Holodiffusion: Training a 3d diffusion model using 2d images.
\newblock In \emph{Proceedings of the IEEE/CVF Conference on Computer Vision and Pattern Recognition}, pages 18423--18433, 2023.

\bibitem[Katzir et~al.(2023)Katzir, Patashnik, Cohen-Or, and Lischinski]{katzir2023noise}
Oren Katzir, Or Patashnik, Daniel Cohen-Or, and Dani Lischinski.
\newblock Noise-free score distillation.
\newblock \emph{arXiv preprint arXiv:2310.17590}, 2023.

\bibitem[Kopf et~al.(2007)Kopf, Fu, Cohen-Or, Deussen, Lischinski, and Wong]{kopf2007solid}
Johannes Kopf, Chi-Wing Fu, Daniel Cohen-Or, Oliver Deussen, Dani Lischinski, and Tien-Tsin Wong.
\newblock Solid texture synthesis from 2d exemplars.
\newblock In \emph{ACM SIGGRAPH 2007 papers}, pages 2--es. 2007.

\bibitem[Kwatra et~al.(2003)Kwatra, Sch{\"o}dl, Essa, Turk, and Bobick]{kwatra2003graphcut}
Vivek Kwatra, Arno Sch{\"o}dl, Irfan Essa, Greg Turk, and Aaron Bobick.
\newblock Graphcut textures: Image and video synthesis using graph cuts.
\newblock \emph{Acm transactions on graphics (tog)}, 22\penalty0 (3):\penalty0 277--286, 2003.

\bibitem[Laine et~al.(2020)Laine, Hellsten, Karras, Seol, Lehtinen, and Aila]{Laine2020diffrast}
Samuli Laine, Janne Hellsten, Tero Karras, Yeongho Seol, Jaakko Lehtinen, and Timo Aila.
\newblock Modular primitives for high-performance differentiable rendering.
\newblock \emph{ACM Transactions on Graphics}, 39\penalty0 (6), 2020.

\bibitem[Lei et~al.(2022)Lei, Zhang, Jia, et~al.]{lei2022tango}
Jiabao Lei, Yabin Zhang, Kui Jia, et~al.
\newblock Tango: Text-driven photorealistic and robust 3d stylization via lighting decomposition.
\newblock \emph{Advances in Neural Information Processing Systems}, 35:\penalty0 30923--30936, 2022.

\bibitem[Levoy et~al.(2000)Levoy, Pulli, Curless, Rusinkiewicz, Koller, Pereira, Ginzton, Anderson, Davis, Ginsberg, et~al.]{levoy2000digital}
Marc Levoy, Kari Pulli, Brian Curless, Szymon Rusinkiewicz, David Koller, Lucas Pereira, Matt Ginzton, Sean Anderson, James Davis, Jeremy Ginsberg, et~al.
\newblock The digital michelangelo project: 3d scanning of large statues.
\newblock In \emph{Proceedings of the 27th annual conference on Computer graphics and interactive techniques}, pages 131--144, 2000.

\bibitem[Li et~al.(2022)Li, Li, Xiong, and Hoi]{li2022blip}
Junnan Li, Dongxu Li, Caiming Xiong, and Steven Hoi.
\newblock Blip: Bootstrapping language-image pre-training for unified vision-language understanding and generation, 2022.

\bibitem[Lin et~al.(2023)Lin, Gao, Tang, Takikawa, Zeng, Huang, Kreis, Fidler, Liu, and Lin]{lin2023magic3d}
Chen-Hsuan Lin, Jun Gao, Luming Tang, Towaki Takikawa, Xiaohui Zeng, Xun Huang, Karsten Kreis, Sanja Fidler, Ming-Yu Liu, and Tsung-Yi Lin.
\newblock Magic3d: High-resolution text-to-3d content creation.
\newblock In \emph{Proceedings of the IEEE/CVF Conference on Computer Vision and Pattern Recognition}, pages 300--309, 2023.

\bibitem[Luo and Hu(2021)]{luo2021diffusion}
Shitong Luo and Wei Hu.
\newblock Diffusion probabilistic models for 3d point cloud generation.
\newblock In \emph{Proceedings of the IEEE/CVF Conference on Computer Vision and Pattern Recognition}, pages 2837--2845, 2021.

\bibitem[Ma et~al.(2023)Ma, Zhang, Sun, Ji, Wang, Jiang, Zhuang, and Ji]{ma2023x}
Yiwei Ma, Xiaoqing Zhang, Xiaoshuai Sun, Jiayi Ji, Haowei Wang, Guannan Jiang, Weilin Zhuang, and Rongrong Ji.
\newblock X-mesh: Towards fast and accurate text-driven 3d stylization via dynamic textual guidance.
\newblock In \emph{Proceedings of the IEEE/CVF International Conference on Computer Vision}, pages 2749--2760, 2023.

\bibitem[Metzer et~al.(2023)Metzer, Richardson, Patashnik, Giryes, and Cohen-Or]{metzer2023latent}
Gal Metzer, Elad Richardson, Or Patashnik, Raja Giryes, and Daniel Cohen-Or.
\newblock Latent-nerf for shape-guided generation of 3d shapes and textures.
\newblock In \emph{Proceedings of the IEEE/CVF Conference on Computer Vision and Pattern Recognition}, pages 12663--12673, 2023.

\bibitem[Michel et~al.(2021)Michel, Bar-On, Liu, Benaim, and Hanocka]{text2mesh}
Oscar Michel, Roi Bar-On, Richard Liu, Sagie Benaim, and Rana Hanocka.
\newblock Text2mesh: Text-driven neural stylization for meshes.
\newblock \emph{arXiv preprint arXiv:2112.03221}, 2021.

\bibitem[Michel et~al.(2022)Michel, Bar-On, Liu, Benaim, and Hanocka]{michel2022text2mesh}
Oscar Michel, Roi Bar-On, Richard Liu, Sagie Benaim, and Rana Hanocka.
\newblock Text2mesh: Text-driven neural stylization for meshes.
\newblock In \emph{Proceedings of the IEEE/CVF Conference on Computer Vision and Pattern Recognition}, pages 13492--13502, 2022.

\bibitem[Mildenhall et~al.(2021)Mildenhall, Srinivasan, Tancik, Barron, Ramamoorthi, and Ng]{mildenhall2021nerf}
Ben Mildenhall, Pratul~P Srinivasan, Matthew Tancik, Jonathan~T Barron, Ravi Ramamoorthi, and Ren Ng.
\newblock Nerf: Representing scenes as neural radiance fields for view synthesis.
\newblock \emph{Communications of the ACM}, 65\penalty0 (1):\penalty0 99--106, 2021.

\bibitem[Mohammad~Khalid et~al.(2022)Mohammad~Khalid, Xie, Belilovsky, and Popa]{mohammad2022clip}
Nasir Mohammad~Khalid, Tianhao Xie, Eugene Belilovsky, and Tiberiu Popa.
\newblock Clip-mesh: Generating textured meshes from text using pretrained image-text models.
\newblock In \emph{SIGGRAPH Asia 2022 conference papers}, pages 1--8, 2022.

\bibitem[M{\"u}ller et~al.(2022)M{\"u}ller, Evans, Schied, and Keller]{muller2022instant}
Thomas M{\"u}ller, Alex Evans, Christoph Schied, and Alexander Keller.
\newblock Instant neural graphics primitives with a multiresolution hash encoding.
\newblock \emph{ACM Transactions on Graphics (ToG)}, 41\penalty0 (4):\penalty0 1--15, 2022.

\bibitem[Munkberg et~al.(2022)Munkberg, Hasselgren, Shen, Gao, Chen, Evans, M{\"u}ller, and Fidler]{munkberg2022extracting}
Jacob Munkberg, Jon Hasselgren, Tianchang Shen, Jun Gao, Wenzheng Chen, Alex Evans, Thomas M{\"u}ller, and Sanja Fidler.
\newblock Extracting triangular 3d models, materials, and lighting from images.
\newblock In \emph{Proceedings of the IEEE/CVF Conference on Computer Vision and Pattern Recognition}, pages 8280--8290, 2022.

\bibitem[Park et~al.(2019)Park, Florence, Straub, Newcombe, and Lovegrove]{park2019deepsdf}
Jeong~Joon Park, Peter Florence, Julian Straub, Richard Newcombe, and Steven Lovegrove.
\newblock Deepsdf: Learning continuous signed distance functions for shape representation.
\newblock In \emph{Proceedings of the IEEE/CVF conference on computer vision and pattern recognition}, pages 165--174, 2019.

\bibitem[Paszke et~al.(2019)Paszke, Gross, Massa, Lerer, Bradbury, Chanan, Killeen, Lin, Gimelshein, Antiga, et~al.]{paszke2019pytorch}
Adam Paszke, Sam Gross, Francisco Massa, Adam Lerer, James Bradbury, Gregory Chanan, Trevor Killeen, Zeming Lin, Natalia Gimelshein, Luca Antiga, et~al.
\newblock Pytorch: An imperative style, high-performance deep learning library.
\newblock \emph{Advances in neural information processing systems}, 32, 2019.

\bibitem[Pavllo et~al.(2021)Pavllo, Kohler, Hofmann, and Lucchi]{pavllo2021learning}
Dario Pavllo, Jonas Kohler, Thomas Hofmann, and Aurelien Lucchi.
\newblock Learning generative models of textured 3d meshes from real-world images.
\newblock In \emph{Proceedings of the IEEE/CVF International Conference on Computer Vision}, pages 13879--13889, 2021.

\bibitem[Poole et~al.(2022)Poole, Jain, Barron, and Mildenhall]{poole2022dreamfusion}
Ben Poole, Ajay Jain, Jonathan~T Barron, and Ben Mildenhall.
\newblock Dreamfusion: Text-to-3d using 2d diffusion.
\newblock \emph{arXiv preprint arXiv:2209.14988}, 2022.

\bibitem[Qian et~al.(2023)Qian, Mai, Hamdi, Ren, Siarohin, Li, Lee, Skorokhodov, Wonka, Tulyakov, et~al.]{qian2023magic123}
Guocheng Qian, Jinjie Mai, Abdullah Hamdi, Jian Ren, Aliaksandr Siarohin, Bing Li, Hsin-Ying Lee, Ivan Skorokhodov, Peter Wonka, Sergey Tulyakov, et~al.
\newblock Magic123: One image to high-quality 3d object generation using both 2d and 3d diffusion priors.
\newblock \emph{arXiv preprint arXiv:2306.17843}, 2023.

\bibitem[Qin et~al.(2020)Qin, Zhang, Huang, Dehghan, Zaiane, and Jagersand]{Qin_2020_PR}
Xuebin Qin, Zichen Zhang, Chenyang Huang, Masood Dehghan, Osmar Zaiane, and Martin Jagersand.
\newblock U2-net: Going deeper with nested u-structure for salient object detection.
\newblock page 107404, 2020.

\bibitem[Radford et~al.(2021)Radford, Kim, Hallacy, Ramesh, Goh, Agarwal, Sastry, Askell, Mishkin, Clark, et~al.]{radford2021learning}
Alec Radford, Jong~Wook Kim, Chris Hallacy, Aditya Ramesh, Gabriel Goh, Sandhini Agarwal, Girish Sastry, Amanda Askell, Pamela Mishkin, Jack Clark, et~al.
\newblock Learning transferable visual models from natural language supervision.
\newblock In \emph{International conference on machine learning}, pages 8748--8763. PMLR, 2021.

\bibitem[Raj et~al.(2023)Raj, Kaza, Poole, Niemeyer, Ruiz, Mildenhall, Zada, Aberman, Rubinstein, Barron, et~al.]{raj2023dreambooth3d}
Amit Raj, Srinivas Kaza, Ben Poole, Michael Niemeyer, Nataniel Ruiz, Ben Mildenhall, Shiran Zada, Kfir Aberman, Michael Rubinstein, Jonathan Barron, et~al.
\newblock Dreambooth3d: Subject-driven text-to-3d generation.
\newblock \emph{arXiv preprint arXiv:2303.13508}, 2023.

\bibitem[Ramesh et~al.(2022)Ramesh, Dhariwal, Nichol, Chu, and Chen]{ramesh2022hierarchical}
Aditya Ramesh, Prafulla Dhariwal, Alex Nichol, Casey Chu, and Mark Chen.
\newblock Hierarchical text-conditional image generation with clip latents.
\newblock \emph{arXiv preprint arXiv:2204.06125}, 1\penalty0 (2):\penalty0 3, 2022.

\bibitem[Richardson et~al.(2023)Richardson, Metzer, Alaluf, Giryes, and Cohen-Or]{richardson2023texture}
Elad Richardson, Gal Metzer, Yuval Alaluf, Raja Giryes, and Daniel Cohen-Or.
\newblock Texture: Text-guided texturing of 3d shapes.
\newblock In \emph{ACM SIGGRAPH 2023 Conference Proceedings}, New York, NY, USA, 2023. Association for Computing Machinery.

\bibitem[Ruiz et~al.(2023)Ruiz, Li, Jampani, Pritch, Rubinstein, and Aberman]{ruiz2023dreambooth}
Nataniel Ruiz, Yuanzhen Li, Varun Jampani, Yael Pritch, Michael Rubinstein, and Kfir Aberman.
\newblock Dreambooth: Fine tuning text-to-image diffusion models for subject-driven generation.
\newblock In \emph{Proceedings of the IEEE/CVF Conference on Computer Vision and Pattern Recognition}, pages 22500--22510, 2023.

\bibitem[Sharma et~al.(2023)Sharma, Jampani, Li, Jia, Lagun, Durand, Freeman, and Matthews]{sharma2023alchemist}
Prafull Sharma, Varun Jampani, Yuanzhen Li, Xuhui Jia, Dmitry Lagun, Fredo Durand, William~T Freeman, and Mark Matthews.
\newblock Alchemist: Parametric control of material properties with diffusion models.
\newblock \emph{arXiv preprint arXiv:2312.02970}, 2023.

\bibitem[Shi et~al.(2023{\natexlab{a}})Shi, Chen, Zhang, Liu, Xu, Wei, Chen, Zeng, and Su]{shi2023zero123plus}
Ruoxi Shi, Hansheng Chen, Zhuoyang Zhang, Minghua Liu, Chao Xu, Xinyue Wei, Linghao Chen, Chong Zeng, and Hao Su.
\newblock Zero123++: a single image to consistent multi-view diffusion base model, 2023{\natexlab{a}}.

\bibitem[Shi et~al.(2023{\natexlab{b}})Shi, Wang, Ye, Long, Li, and Yang]{shi2023mvdream}
Yichun Shi, Peng Wang, Jianglong Ye, Mai Long, Kejie Li, and Xiao Yang.
\newblock Mvdream: Multi-view diffusion for 3d generation.
\newblock \emph{arXiv preprint arXiv:2308.16512}, 2023{\natexlab{b}}.

\bibitem[Siddiqui et~al.(2022)Siddiqui, Thies, Ma, Shan, Nie{\ss}ner, and Dai]{siddiqui2022texturify}
Yawar Siddiqui, Justus Thies, Fangchang Ma, Qi Shan, Matthias Nie{\ss}ner, and Angela Dai.
\newblock Texturify: Generating textures on 3d shape surfaces.
\newblock In \emph{European Conference on Computer Vision}, pages 72--88. Springer, 2022.

\bibitem[Sohl-Dickstein et~al.(2015)Sohl-Dickstein, Weiss, Maheswaranathan, and Ganguli]{sohl2015deep}
Jascha Sohl-Dickstein, Eric Weiss, Niru Maheswaranathan, and Surya Ganguli.
\newblock Deep unsupervised learning using nonequilibrium thermodynamics.
\newblock In \emph{International conference on machine learning}, pages 2256--2265. PMLR, 2015.

\bibitem[Song et~al.(2020)Song, Sohl-Dickstein, Kingma, Kumar, Ermon, and Poole]{song2020score}
Yang Song, Jascha Sohl-Dickstein, Diederik~P Kingma, Abhishek Kumar, Stefano Ermon, and Ben Poole.
\newblock Score-based generative modeling through stochastic differential equations.
\newblock \emph{arXiv preprint arXiv:2011.13456}, 2020.

\bibitem[Sun et~al.(2023)Sun, Cai, Li, Yan, Zhang, Marshall, Huang, Zhao, and Dong]{sun2023neural}
Cheng Sun, Guangyan Cai, Zhengqin Li, Kai Yan, Cheng Zhang, Carl Marshall, Jia-Bin Huang, Shuang Zhao, and Zhao Dong.
\newblock Neural-pbir reconstruction of shape, material, and illumination.
\newblock In \emph{Proceedings of the IEEE/CVF International Conference on Computer Vision (ICCV)}, 2023.

\bibitem[Wang et~al.(2023{\natexlab{a}})Wang, Du, Li, Yeh, and Shakhnarovich]{wang2023score}
Haochen Wang, Xiaodan Du, Jiahao Li, Raymond~A Yeh, and Greg Shakhnarovich.
\newblock Score jacobian chaining: Lifting pretrained 2d diffusion models for 3d generation.
\newblock In \emph{Proceedings of the IEEE/CVF Conference on Computer Vision and Pattern Recognition}, pages 12619--12629, 2023{\natexlab{a}}.

\bibitem[Wang et~al.(2023{\natexlab{b}})Wang, Lu, Wang, Bao, Li, Su, and Zhu]{wang2023prolificdreamer}
Zhengyi Wang, Cheng Lu, Yikai Wang, Fan Bao, Chongxuan Li, Hang Su, and Jun Zhu.
\newblock Prolificdreamer: High-fidelity and diverse text-to-3d generation with variational score distillation.
\newblock \emph{arXiv preprint arXiv:2305.16213}, 2023{\natexlab{b}}.

\bibitem[Wu et~al.(2016)Wu, Zhang, Xue, Freeman, and Tenenbaum]{wu2016learning}
Jiajun Wu, Chengkai Zhang, Tianfan Xue, Bill Freeman, and Josh Tenenbaum.
\newblock Learning a probabilistic latent space of object shapes via 3d generative-adversarial modeling.
\newblock \emph{Advances in neural information processing systems}, 29, 2016.

\bibitem[Yan et~al.(2023)Yan, Luan, Ha\v{s}an, Groueix, Deschaintre, and Zhao]{Yan:2023:PSDR-Room}
K. Yan, F. Luan, M. Ha\v{s}an, T. Groueix, V. Deschaintre, and S. Zhao.
\newblock Psdr-room: Single photo to scene using differentiable rendering.
\newblock In \emph{ACM SIGGRAPH Asia 2023 Conference Proceedings}, 2023.

\bibitem[Yu et~al.(2023)Yu, Dai, Li, Ma, Liu, and Qi]{yu2023texture}
Xin Yu, Peng Dai, Wenbo Li, Lan Ma, Zhengzhe Liu, and Xiaojuan Qi.
\newblock Texture generation on 3d meshes with point-uv diffusion.
\newblock In \emph{Proceedings of the IEEE/CVF International Conference on Computer Vision}, pages 4206--4216, 2023.

\bibitem[Zhang et~al.(2023)Zhang, Rao, and Agrawala]{zhang2023adding}
Lvmin Zhang, Anyi Rao, and Maneesh Agrawala.
\newblock Adding conditional control to text-to-image diffusion models.
\newblock In \emph{Proceedings of the IEEE/CVF International Conference on Computer Vision}, pages 3836--3847, 2023.

\bibitem[Zhou and Koltun(2014)]{zhou2014color}
Qian-Yi Zhou and Vladlen Koltun.
\newblock Color map optimization for 3d reconstruction with consumer depth cameras.
\newblock \emph{ACM Transactions on Graphics (ToG)}, 33\penalty0 (4):\penalty0 1--10, 2014.

\end{thebibliography}
}

\end{document}